\documentclass[10pt,conference]{IEEEtran}
\IEEEoverridecommandlockouts
\usepackage{cite}
\usepackage{amsmath,amssymb,amsfonts}
\usepackage{algorithmic}
\usepackage{graphicx}
\usepackage{textcomp}
\usepackage{xcolor}
\usepackage{caption}
\usepackage{subcaption}
\usepackage{booktabs}

\def\BibTeX{{\rm B\kern-.05em{\sc i\kern-.025em b}\kern-.08em
    T\kern-.1667em\lower.7ex\hbox{E}\kern-.125emX}}

\begin{document}

\title{Revisiting Deep Learning for \\Variable Type Recovery}

\author{\IEEEauthorblockN{Kevin Cao}
\IEEEauthorblockA{\textit{Vanderbilt University} \\
Nashville, USA \\
\texttt{kevin.cao@vanderbilt.edu}}
\and
\IEEEauthorblockN{Kevin Leach}
\IEEEauthorblockA{\textit{Vanderbilt University} \\
Nashville, USA \\
\texttt{kevin.leach@vanderbilt.edu}}
}

\maketitle

\begin{abstract}
Compiled binary executables are often the only available artifact in reverse engineering, malware analysis, and software systems maintenance. 
Unfortunately, the lack of semantic information like variable types makes comprehending binaries difficult. 
In efforts to improve the comprehensibility of binaries, researchers have recently used machine learning techniques to predict semantic information contained in the original source code.
Chen et al. implemented DIRTY, a Transformer-based Encoder-Decoder architecture capable of augmenting decompiled code with variable names and types by leveraging decompiler output tokens and variable size information.
Chen et al. were able to demonstrate a substantial increase in name and type extraction accuracy on Hex-Rays decompiler outputs compared to existing static analysis and AI-based techniques.
We extend the original DIRTY results by re-training the DIRTY model on a dataset produced by the open-source Ghidra decompiler.
Although Chen et al. concluded that Ghidra was not a suitable decompiler candidate due to its difficulty in parsing and incorporating DWARF symbols during analysis, we demonstrate that straightforward parsing of variable data generated by Ghidra results in similar retyping performance.
We hope this work inspires further interest and adoption of the Ghidra decompiler for use in research projects.
\end{abstract}

\begin{IEEEkeywords}
Ghidra, Hex-Rays, Machine Learning, Transformers
\end{IEEEkeywords}

\section{Introduction}
Program comprehension is a critical part of developing and maintaining large software systems. 
Many analysis and comprehension tools operate on program source code, such as code similarity comparison~\cite{SIGMOD-2003-SchleimerWA}, automated program repair~\cite{LeGoues12tse}, and fault localization~\cite{PearsonCJFAEPK2017}.
However, when dealing with legacy systems or proprietary software, researchers and reverse engineers often have access only to the distributed compiled binaries.
Binary files are frequently viewed as more difficult to analyze than source code files, in part because assembly and machine language lack the semantics and structure present in a higher-level programming language.
Although debugging formats such as DWARF allow compilers to build binaries with debugging symbols to integrate semantic information within the binary, release builds often omit these symbols to improve performance and conceal intellectual property.

To analyze binary files, reverse engineers use decompilers to transform a binary file (or single function) into an equivalent source code representation, which in turn helps engineers comprehend the original semantics of the file or function in question.
This transformation augments the sequential, type-agnostic nature of assembly programming with abstractions such as variables names and types.
However, without the original semantic information, decompilers are unable to provide meaningful names to these variables, instead assigning names and types based on implementation-specific naming conventions (e.g., `uVar1').

With the success of machine learning models in the domains of natural language processing~\cite{brown2020language} and programming language analysis~\cite{feng2020codebert}, researchers are proposing machine learning models to recover missing semantic information in binaries.
These models leverage the insight that semantic information present in source code is context-dependent: variables that appear and are used in similar contexts tend to be assigned similar names and types~\cite{7961504}.
One such approach is the DIRTY (\textbf{D}ecomp\textbf{I}led variable \textbf{R}e\textbf{TY}per) model proposed by Chen et al.~\cite{chen2021augmenting,Lacomis2019dire}, which adapts the Transformer architecture to predict variable names and types in decompiler outputs.
Due to its focus on variable types, DIRTY leverages both the decompiled source code tokens as well as the object layout for all variables found during decompilation
to improve model performance.
This architecture, combined with the novel use of variable layouts, resulted in a model capable of identifying correct variable types 75.8\% of the time. 

One main concern with the DIRTY model is its ability to generalize to different decompiler outputs, as a swath of tools are commonly used in reverse engineering~\cite{binja,remill,ghidra,7546500}. 
The authors tested their implementation solely using the commercial Hex-Rays decompiler, whose expensive licensing restrictions are outside the budget of many researchers and hobbyist reverse engineers.
For this reproducibility study, we address this concern by training the DIRTY architecture with a dataset decompiled using the open-source Ghidra decompiler.
We observe that the performance of this neural architecture is comparable for both Hex-Rays decompiler outputs and Ghidra decompiler outputs.

The use of Ghidra is addressed in the DIRTY paper, where the authors concluded that the inability of Ghidra to reliably obtain correct ground truth semantic information from debugging symbols in the data processing stage harmed the performance of the model.
While we do not address specific decompilation algorithms in this paper, we adapt the dataset used in DIRTY by including only those binaries decompiled by Ghidra that contain corresponding DWARF symbolic information that can serve as adequate ground truth. 
We show that this dataset permits the DIRTY architecture to generalize to decompiler output produced with Ghidra for variable naming and typing tasks. 
We hope that this reproducibility study will aid future researchers seeking to incorporate Ghidra and other reverse engineering tools into datasets and evaluations. 


In summary, we demonstrate that DIRTY's architecture for the task of variable type prediction generalizes to our newly curated Ghidra dataset.

\section{Related Work}

Binaries have been an important artifact for applying techniques to aid reverse engineering tasks.
While machine learning models exist that analyze binary files directly, these models often target specific tasks such as optimization level recovery that leverage insights on broad binary characteristics such as instruction frequency \cite{9825843}.
Models like DIRTY attempt to provide reverse engineers with additional semantics to augment source code produced by decompilers, ultimately improving comprehensibility. 

Like other machine learning techniques tackling binary analyses such as binary similarity~\cite{binshot}, vulnerability detection~\cite{giffin2004efficient}, and malware analysis~\cite{mimosa}, DIRTY trains on decompiled functions.
Informally, constructing the dataset for DIRTY involves decompiling a set of binaries for which DWARF symbols are available, resulting in tuples of (assembly code, variable names), which in turn forms ground truth that can be used for training.
This is possible because the decompiler interface is scriptable, allowing programmatic access to the internal abstract syntax tree, object layouts, and underlying type system to transform the decompiled code produced by the base set of decompilation passes.
These features exist in all decompiler implementations because they are fundamental to the construction of higher level languages.
Therefore, researchers can leverage these features to build general machine learning models that eliminate implementation-specific details.

The original DIRTY architecture was trained solely using the Hex-Rays decompiler.
Despite the rich integration supported by Hex-Rays, other tools are frequently used in reverse engineering and binary comprehension tasks. 
The Hex-Rays decompiler is a closed-source software product entailing licenses specific to one backend CPU target (e.g., x86, ARM on Windows, Linux), which limits augmentation and widespread adoption.
This can pose substantial barriers for those seeking to use Hex-Rays in larger, more complex machine learning techniques that depend on large volumes of decompiled code, precluding independent researchers and hobbyists from advancing the field. 

Many decompiler frameworks have been released to address these concerns. 
In this paper, we choose to augment DIRTY by retraining the neural architecture on a dataset derived from Ghidra\footnote{Ghidra: https://github.com/NationalSecurityAgency/ghidra}, a public version of the decompiler used by the National Security Agency released in 2019.
Due in part to its recent release, features such as native scripting capabilities and DWARF parsing lag behind their Hex-Rays counterparts. 
However, adapting binary analysis techniques to work with Ghidra is important to ensure generalizable scientific findings that increase accessibility to important tools that advance the state-of-the-art in binary analysis and decompilation.
Several other reverse engineering frameworks exist such as angr\cite{7546500}, a framework that goes beyond decompilation and disassembly by introducing features such as symbolic execution, Radare2\footnote{Radare2: https://github.com/radareorg/radare2}, a framework cenetered around disassembly that offers interfacing with external decompilers, and Binary Ninja~\cite{binja}, a framework focusing on UI usability and automation while still offering the same scripting capabilities as other frameworks.
We ultimately select Ghidra due to its similar capabilities to Hex-Rays and community interest in seeing Ghidra-based decompilation projects.

\section{Experimental Design}
\subsection{Data Processing and Training}
\begin{figure}[htbp]
     \centering
    \includegraphics{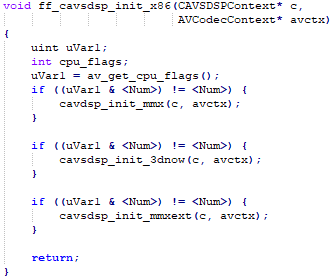}
    \caption{Example of aliasing in Ghidra. While Ghidra finds the cpu\_flags variable in the DWARF section, Ghidra assigns flag information to an unnamed uVar1 variable. These two variables are difficult to programmatically combine into one variable since they have different memory locations. }
    \label{fig:ghidra_dwarf_ex}
\end{figure}

In this section, we introduce the experimental design for our replication study.
First, we introduce the DIRT dataset, which serves as the basis for the dataset we curated to train our Ghidra-specific DIRTY model.
We then introduce the processing, training, and testing of the model while highlighting differences between the original implementation when necessary.
For future reference, we refer to DIRTY\textsubscript{Hex-Rays} as the original model implementation in the DIRTY paper and DIRTY\textsubscript{Ghidra} as our model trained on Ghidra decompiled functions.

\subsection{DIRT and Dataset Creation}
The \textbf{D}ataset for \textbf{I}diomatic \textbf{R}e\textbf{T}yping is a dataset created from randomly sampling C and C++ projects scraped from GitHub using GHCC \footnote{GHCC: https://github.com/huzecong/ghcc}, which attempts to compile each project with debugging symbols by recursively identifying Makefiles in the project's sub-directories.
Each binary file is decompiled using Hex-Rays to obtain ground truth information about variable names and layouts.
The binary files are then stripped of their debugging symbols and decompiled once again to serve as indicative outputs obtained from real stripped binaries in the wild. 
The final \textbf{DIRT} dataset consists of a collection of decompiled code tokens with variables clearly marked as well as object size (e.g., struct size) information obtained from the decompiler for all variables in the function.

We are interested in training the DIRTY model using output from the Ghidra decompiler.
While we could compile our own set of binaries to create our dataset, assembling enough binaries to create a dataset comparable in size with the DIRT dataset is a large undertaking.
More importantly, we fear that changes in compilation environments such as compiler versions might undermine the ability to directly compare model results.
Although only the processed versions of the DIRT testing and training sets are publicly available, Chen et al. were kind enough to provide the underlying collection of binaries used to select the DIRT binaries.
Using this collection of binaries, we created our dataset by combining the original binaries used in the DIRT dataset along with an additional set of 90,000 randomly sampled binaries to ensure we had a dataset size of a similar order of magnitude after preprocessing.
We decompile these binaries using Ghidra and extract the C-code functions and object layouts using Ghidra's underlying decompiler interface.

As seen in Fig.~\ref{fig:ghidra_dwarf_ex}, we observed that the Ghidra decompiler suffers from a form of variable aliasing when decompiling binaries with debugging information.
While Ghidra recovers the \texttt{cpu\_flags} DWARF variable and includes it in the decompiled function source, Ghidra fails to assign values to this variable, instead assigning \texttt{cpu\_flags} information to a decompiler-created \texttt{uVar1} variable, which does not contain any debugging information.
Although reverse engineers reading this function might be able to see that \texttt{uVar1} aliases the \texttt{cpu\_flags} variable and thus combine the two variables to create a cleaner, more accurate decompiled function, these two variables do not reside in the same memory location in Ghidra, which prevents us from confidently merging these two variables during preprocessing.
It becomes more impractical to implement variable merging when multiple variables are being aliased: the correct mapping between variables is now context-dependent.
This seems to be the issue that Chen et al. faced when trying to reproduce their own work using Ghidra.

Although resolving variable aliasing is an important undertaking requiring substantial reverse engineering expertise,
we identify two insights that allow us to work around this problem.
First, the concept of aliasing is not specific to the Ghidra decompiler: typical decompilers face this issue since binaries lack rich semantics that are lost during compilation.
Indeed, we see that, for some functions, Ghidra correctly uses variables found in DWARF debugging sections, and Hex-Rays suffers from a similar problem in variables suddenly appearing when decompiling stripped binaries compared to binaries with debugging information.
Secondly, we note that this inability to correctly assign variables with DWARF types actually conveys semantic information, since it signifies that the variable is either a temporary or was combined with other variables in the source code during analysis.

Thus, we do not remove such ``bad'' DWARF variables from the training set because they are variables encountered by Ghidra during analysis, and we do not want to artificially bias the results of our replicated model by introducing a bad data exclusion step not present in the original published results.
Therefore, we annotate these variables with a special \emph{disappear} label to signify that DWARF information was lost during decompilation.
Like \emph{primitive}, \emph{pointer}, and \emph{structure} data types, this \emph{disappear} type can be predicted by the model during retyping tasks.

To the best of our knowledge, Ghidra does not provide an interface to determine whether or not a variable contains DWARF information.
Instead, before decompilation, we manually parse the DWARF information ourselves to extract the DWARF names of all variables declared in the binary.
Thus, during decompilation, if a variable is found that does not match a name discovered during our DWARF parsing stage, we replace its type with our \emph{disappear} datatype.

We implement various other techniques to filter the functions decompiled by Ghidra, summarized as follows.

\begin{enumerate}
    \item We enforce a decompilation time limit of 3 minutes to prevent the decompiler from stalling on large executables.
    \item We exclude functions that Ghidra fails to decompile cleanly (e.g., external functions) to prevent the model from training on incomplete function signatures and bodies.
    \item We exclude functions that do not reference parameters or contain local variables in the function body as they are not useful for the variable naming task.
\end{enumerate}

The breakdown between the DIRT original dataset and our curated version of the dataset is shown in Table~\ref{TrainDatasets} and Table.~\ref{TestDatasets}.
We see that variable aliasing issues are more prevalent in Ghidra, where 62.15\% of Ghidra variables lack debugging information compared to the 20.13\% in Hex-Rays.

\begin{table*}[htbp]
    \centering
    \setlength{\tabcolsep}{1.3mm}{
    \caption{Comparison between the datasets used to train DIRTY\textsubscript{Hex-Rays} and DIRTY\textsubscript{Ghidra}. Due to the high number of \emph{disappear} variables and low number of \emph{struct} variables, we address these types separately from others during evaluation.}\label{TrainDatasets}
    \begin{tabular}{ l  r r r r  r }
        \toprule
        &  & \multicolumn{3}{c}{Variables} & \\
        Model & \# Binaries & \# Variables & \% Structs & \% Disappear & \# Unique Functions  \\
        \midrule
        DIRTY\textsubscript{Hex-Rays} & 76,472 & 3,689,018 & 5.05 & 20.13 & 727,617\\
        \midrule
        DIRTY\textsubscript{Ghidra} & 73,232 & 1,676,346 & 2.45 & 62.15 & 399,130 \\
        \bottomrule
    \end{tabular}}
\end{table*}

\begin{table*}[htbp]
    \centering
    \setlength{\tabcolsep}{1.3mm}{
    \caption{Comparison between the datasets used to evaluate DIRTY\textsubscript{Hex-Rays} and DIRTY\textsubscript{Ghidra}. A lower In-Train percentage for the Ghidra dataset signifies less shared library functions in the Ghidra testing and training sets.}\label{TestDatasets}
    \begin{tabular}{ l  r  r r r  r r r  r }
        \toprule
        &  & \multicolumn{3}{c}{Variables} & \multicolumn{3}{c}{Functions} & \\
        Model & \# Binaries & \# Variables & \% Structs & \% Disappear & \# Functions & \% All Disappear & \% No disappear & \% In-Train  \\
        \midrule
        DIRTY\textsubscript{Hex-Rays} & 9,461 & 1,031,844 & 1.83 & 8.76 & 203,876 & 0.87 & 74.84 & 65.5\\
        \midrule
        DIRTY\textsubscript{Ghidra} & 9,153 & 307,083 & 1.22 & 42.78 & 86,870 & 26.1 & 34.71 & 45.04\\
        \bottomrule
    \end{tabular}}
\end{table*}

\begin{table*}[htbp]
    \centering
    \setlength{\tabcolsep}{1.3mm}{
    \caption{Comparison between the variable typing accuracies of the two models, separated by structs, variables labeled \emph{disappear}, and recovered variables. While the Not In-Train accuracies for Hex-Rays\textsubscript{Ghidra} might be slightly higher, a lower \% In-Train causes the Overall accuracies to be similar. }\label{AccTable}
    \begin{tabular}{ l  r r r r  r r r r  r r r r }
        \toprule
        & \multicolumn{4}{c}{Overall} & \multicolumn{4}{c}{In-Train} & \multicolumn{4}{c}{Not In-Train}\\
        Model & Overall & Structs & Disappear & No Disappear & Overall & Structs & Disappear & No Disappear & Overall & Structs & Disappear & No Disappear \\
        \midrule
        DIRTY\textsubscript{Hex-Rays} & 73.9 & 65.8 & 84.7 & 72.8 & 88.1 & 76.6 & 93.4 & 87.5 & 54.3 & 51.7 & 71.5 & 52.8 \\
        \midrule
        DIRTY\textsubscript{Ghidra} & 73.5 & 61.7 & 84.7 & 65.2 & 91.2 & 90.7 & 93.1 & 89.5 & 64.1 & 57.1 & 79.6 & 53.2 \\
        \bottomrule
    \end{tabular}}
\end{table*}

\subsection{Evaluation}
To evaluate the predictions made by the Transformer model, we employ the same metrics as in the DIRTY paper.
A variable type prediction is correct if and only if the type name, along with all other sub-fields if dealing with structure types, are equivalent to the developer-assigned type (as defined in the original source code) using direct string comparison.
While this evaluation metric is strict as incorrect names can still provide important semantic information, string comparison is both simple to implement and deterministic.
By keeping the evaluation metrics the same, we ensure a fair comparison in the final results since any changes in model accuracy can be attributed to differences in the decompiler outputs of Ghidra and Hex-Rays instead of nuances in the evaluation scheme.

\section{Results}

In this section, we aim to answer the following research question: Is the Transformer architecture employed in DIRTY as effective at predicting variable names and types for a dataset decompiled with Ghidra instead of Hex-Rays?

Recall that we retrained the neural architecture in DIRTY using our Ghidra-decompiled dataset.  
We obtained a pre-trained DIRTY model, then evaluated our trained Ghidra model compared to their pre-trained Hex-Rays model on the task for variable retyping.
The results are shown in Table~\ref{AccTable}.
Due to the low amount of structure variables and high levels of \emph{disappear} variables in the Ghidra dataset, we isolate these variable types and obtain their accuracies separately.

We see that the overall performance of the two models have similar overall accuracy, with DIRTY\textsubscript{Hex-Rays} retyping variables correct 73.9\% of the time and DIRTY\textsubscript{Ghidra} retyping variables correct 73.5\% of the time.
While DIRTY\textsubscript{Hex-Rays} and DIRTY\textsubscript{Ghidra} have similar accuracies for retyping variables in functions encountered during training, DIRTY\textsubscript{Ghidra} performs slightly better than DIRTY\textsubscript{Hex-Rays} when retyping variables in functions not seen during training.

For variables without corresponding DWARF variable information, DIRTY\textsubscript{Ghidra} outperforms DIRTY\textsubscript{Hex-Rays} by 5\%.
While the increase in prediction rates can be attributed to the increased proportion of these variables in the Ghidra dataset, the high rate of accurate predictions suggests that for both the Hex-Rays and Ghidra decompilers, there are certain underlying patterns that unify cases in which DWARF semantic information is unrecoverable.

For variables with corresponding DWARF variable information, DIRTY\textsubscript{Ghidra} performs similarly to DIRTY\textsubscript{Hex-Rays}, correctly predicting the DWARF types 53.2\% and 52.8\% of the time respectively.
This result shows that the Ghidra model's ability to predict types was not adversely affected by the greater number of \emph{disappear} variables in the Ghidra dataset.

Overall, these results match our intuition and provide additional evidence that the model described in DIRTY can also apply to decompiled source code obtained from Ghidra.

\section{Conclusion}
In this paper, we extend the work of Chen et al. by demonstrating that for their DIRTY architecture, a model trained on a Ghidra dataset yields similar retyping accuracies as the original model trained on a Hex-Rays dataset.
The model trained on Ghidra decompiler output is capable of correctly predicting DWARF debugging types 73.5\% of the time and displays similar accuracies for function encountered during training and novel functions.
Our results add confidence to the claim that DIRTY's Transformer architecture extends to different decompilers besides Hex-Rays.
We hope that our initial results spark more confidence and adoption of the Ghidra decompiler in related research projects.

\section{Acnkowledgments}

We acknowledge Jeremy Lacomis and Claire Le Goues from the DIRTY paper for graciously providing the original dataset of binaries and scripts used to train their published model.  

\bibliographystyle{IEEEtran}
\bibliography{references}

\end{document}